\documentclass[conference]{IEEEtran}
\IEEEoverridecommandlockouts
\usepackage{cite}
\usepackage{amsmath,amssymb,amsfonts}
\usepackage{graphicx}
\usepackage{array}
\usepackage{float}
\usepackage{comment}

\def\BibTeX{{\rm B\kern-.05em{\sc i\kern-.025em b}\kern-.08em
    T\kern-.1667em\lower.7ex\hbox{E}\kern-.125emX}}

\usepackage{tikz}
\usepackage{hyperref}
\usepackage{lipsum}

\newcommand\copyrighttext{%
  \footnotesize \textcopyright 2025 IEEE. Personal use of this material is permitted.
  Permission from IEEE must be obtained for all other uses, in any current or future 
  media, including reprinting/republishing this material for advertising or promotional 
  purposes, creating new collective works, for resale or redistribution to servers or 
  lists, or reuse of any copyrighted component of this work in other works. The definitive version was published in IEEE International Smart Cities Conference (ISC2) Prooceedings, DOI:  \href{https://doi.org/10.1109/ISC266238.2025.11293252}{10.1109/ISC266238.2025.11293287}.
  }
\newcommand\copyrightnotice{%
\begin{tikzpicture}[remember picture,overlay]
\node[anchor=south,yshift=10pt] at (current page.south) {\fbox{\parbox{\dimexpr\textwidth-\fboxsep-\fboxrule\relax}{\copyrighttext}}};
\end{tikzpicture}%
}

\begin{document}

\title{SUSTAINABLE Platform: Seamless Smart Farming Integration Towards Agronomy Automation\\
\thanks{This research was funded by the European Union’s Horizon 2020 research and innovation program under the Marie Sklodowska-Curie-RISE, Project SUSTAINABLE with grant number 101007702. The paper reflects the authors’ views, and the Commission is not responsible for any use that may be made of the information it contains.}
}
\author{
    \IEEEauthorblockN{ 
    Agorakis Bompotas\IEEEauthorrefmark{1},
    Konstantinos Koutras\IEEEauthorrefmark{1},
    Nikitas Rigas Kalogeropoulos\IEEEauthorrefmark{1},\\
    Panagiotis Kechagias\IEEEauthorrefmark{1},
    Dimitra Gariza\IEEEauthorrefmark{1},
    Athanasios P. Kalogeras\IEEEauthorrefmark{1},
    Christos Alexakos\IEEEauthorrefmark{1}}
    \\
    \IEEEauthorblockA{\IEEEauthorrefmark{1}\textit{Industrial Systems Institute} \\
    \textit{ATHENA Research Center}\\
    Patras, Greece \\
    \{abompotas, koutrask, nkalogeropoulos, kechagias, dgariza, kalogeras, alexakos\}@athenarc.gr}

}

\maketitle

\begin{abstract}
The global agricultural sector is undergoing a transformative shift, driven by increasing food demands, climate variability and the need for sustainable practices. SUSTAINABLE is a smart farming platform designed to integrate IoT, AI, satellite imaging, and role-based task orchestration to enable efficient, traceable, and sustainable agriculture with a pilot usecase in viticulture. This paper explores current smart agriculture solutions, presents a comparative evaluation, and introduces SUSTAINABLE's key features, including satellite index integration, real-time environmental data, and role-aware task management tailored to Mediterranean vineyards.
\end{abstract}

\copyrightnotice


\section{Introduction}
The global agricultural sector faces mounting challenges \cite{vos2019global}, including rising food demand \cite{falcon2022rethinking}, climate variability \cite{bhatti2024global}, and increasing concerns over food safety and security \cite{barrett2021overcoming}. Traditional farming methods often struggle to ensure consistent crop quality while minimizing resource waste and contamination risks. In response, the integration of Agriculture 4.0 \cite{rose2018agriculture}—and especially the application of Industry 4.0 technologies in farming \cite{batra2024industrial}—has emerged as a transformative approach to achieving smart, automated and sustainable agronomy. By leveraging advancements such as the Internet of Things (IoT) \cite{ojha2021internet}, artificial intelligence (AI) \cite{ahmad2024ai}, robotics \cite{yepez2023mobile}, and big data analytics \cite{chergui2022data}, modern farming systems can enhance precision, efficiency, and traceability across the agricultural value chain. Different challenges persist in the use and integration of diverse technologies towards  Agriculture 4.0 paradigm \cite{alexopoulos2023complementary}.

Food safety remains a critical issue, with contamination, pesticide misuse, and supply chain inefficiencies posing significant risks to consumer health. Conventional monitoring methods are often reactive, labor-intensive, and prone to human error. However, smart farming platforms enable real-time data collection and automated decision-making, ensuring optimal growing conditions, early pest detection, and compliance with safety standards. For instance, IoT-enabled sensors monitor soil health, moisture, and environmental factors, while AI-driven analytics predict disease outbreaks and optimize inputs. Such automation minimizes human intervention, reducing contamination risks and improving yield consistency.

The present work presents H2020 SUSTAINABLE project \cite{sustainable}, and its agricultural process management platform that actively engages farming ecosystem stakeholders – including agronomists, farmers, and cooperatives – in executing data-informed actions across the entire cultivation cycle, from planting to harvest. Unlike conventional systems that merely aggregate raw data, the platform operates at the information level, transforming inputs from IoT sensors, drone and satellite imagery, smart lab analyzers, and meteorological sources into actionable insights for stakeholders. By integrating ready-to-use solutions not as isolated data streams but as interconnected decision-support tools, the platform enables farmers and agronomists to plan, monitor, and adjust agricultural practices in real time. This approach ensures that stakeholders remain central to the process, leveraging technology to enhance precision, sustainability, and compliance without replacing human expertise.

The rest of the paper is structured as follows. Section II presents related work and smart agriculture approaches, section III details the SUSTAINABLE platform, while section IV presents relevant usecases. Section V provides discussion and conclusions.

\section{Related Work - Integration of Smart Agriculture Technologies}

Several commercial and research-based platforms have emerged to integrate smart technologies in agriculture. Platforms like \textit{xFarm}, \textit{CropX}, \textit{Climate FieldView}, and \textit{OneSoil} provide tools for agronomic data management, vegetation monitoring and analytics to support precision farming. A common yet powerful indicator used in most of these platforms is the Normalized Difference Vegetation Index (NDVI), a remote sensing metric used to quantify vegetation health and density by calculating the difference between reflected near-infrared and red light.

\textit{xFarm} combines weather forecasts with  visualization to help monitor field conditions and plan treatments  \cite{xfarm}. \textit{CropX} uses soil sensors and Machine Learning (ML) based recommendations for irrigation and fertilization \cite{cropx}. \textit{Climate FieldView} focuses on combining satellite and ground-based data to track vegetation variability across the season \cite{fieldview}. \textit{OneSoil} provides real-time NDVI analysis and zoning for input optimization \cite{onesoil}.

Academic approaches complement these platforms. Kulatunga et al. use NDVI and topographic data for ML-driven zoning in vineyards \cite{kulatunga2024}, while Mazzia et al. enhance NDVI accuracy using UAV imagery and deep learning models tailored to viticulture \cite{mazzia2020}. Kumar et al. \cite{kumar2024} review smart agriculture frameworks with emphasis on IoT-enabled data integration for sustainable practices.

These platforms and studies support various agronomic processes at different levels of technological integration. Tables ~\ref{tab:comparison} and ~\ref{tab:comparison-2} present a comparison of commercial platforms and research efforts based on six criteria: vegetation indices, soil/IoT integration, environmental sensing, task/role management, regional specialization, and ML-based zoning. Unlike other projects in this field, SUSTAINABLE is expressly designed for the comprehensive integration of smart agriculture technologies, giving it a distinct advantage.

\begin{table}[H]
\centering
\caption{Comparative Functionality of Selected Platforms and Research Approaches (I)}
\resizebox{\columnwidth}{!}{%
\begin{tabular}{|p{3.5cm}|c|c|c|}
\hline
\textbf{System / Study} & \textbf{NDVI / Indices} & \textbf{Soil / IoT} & \textbf{Env. Sensors} \\
\hline
\textit{xFarm} & \checkmark{} NDVI & Partial & -- \\
\hline
\textit{CropX} & -- & \checkmark{} & -- \\
\hline
\textit{Climate FieldView} & \checkmark{} & -- & -- \\
\hline
\textit{OneSoil} & \checkmark{} real-time & -- & -- \\
\hline
Kulatunga et al. & \checkmark{} + elevation & -- & -- \\
\hline
Mazzia et al. & \checkmark{} refined (UAV+DL) & -- & -- \\
\hline
\textbf{SUSTAINABLE} & \checkmark{} (NDVI, NDMI, OSAVI) & Partial & \checkmark{} \\
\hline
\end{tabular}
}
\label{tab:comparison}
\end{table}

\begin{table}[H]
\centering
\caption{Comparative Functionality of Selected Platforms and Research Approaches (II)}
\resizebox{\columnwidth}{!}{%
\begin{tabular}{|p{3.5cm}|c|c|c|}
\hline
\textbf{System / Study} & \textbf{Task Mgmt / Roles} & \textbf{Regional Focus} & \textbf{ML / Zoning} \\
\hline
\textit{xFarm} & Basic & General & -- \\
\hline
\textit{CropX} & Analytics-only & General & \checkmark{} \\
\hline
\textit{Climate FieldView} & Limited & General & \checkmark{} \\
\hline
\textit{OneSoil} & -- & General & \checkmark{} \\
\hline
Kulatunga et al. & -- & Research & \checkmark{} \\
\hline
Mazzia et al. & -- & Vineyard-focused & \checkmark{} \\
\hline
\textbf{SUSTAINABLE} & \checkmark{} (role-based) & Mediterranean & -- \\
\hline
\end{tabular}
}
\label{tab:comparison-2}
\end{table}

\section{SUSTAINABLE Platform}

\textbf{SUSTAINABLE} platform is a comprehensive, smart agriculture integration platform pilot tested in Mediterranean viticulture and small-to-medium-sized vineyards. It integrates satellite-based vegetation analytics, real-time environmental sensing, and role-driven task coordination to enable sustainable decision-making in modern agriculture.

The platform integrates diverse agriculture support systems, such as platforms analyzing high-resolution satellite imagery with multi-index analysis (Normalized Difference Vegetation Index - NDVI, Normalized Difference Moisture Index - NDMI, Optimized Soil Adjusted Vegetation Index - OSAVI) or even multi-spectral images from drones. Furthermore, environmental conditions are monitored via weather stations providing live data on temperature, humidity, and dew point.

SUSTAINABLE Platform is based on a task management system assisting stakeholders to track and execute field activities (e.g., spraying, harvesting). It integrates external Smart Agriculture Solutions (IoT-based Sensing, Field Image Data Analysis, Automated Irrigation Systems, etc. ) via web-based APIs interacting on exchanging data and providing command.

The main outcome of the SUSTAINABLE Platform is a holistic approach to farm management. It offers users extended management and monitoring capabilities of the relevant processes, like field inspection and irrigation, crop spraying, agronomic assessment, etc. It offers involved actors a single focal point of reference, providing seamless access to the different information sources and systems utilized in their everyday routine. It stands up to the challenge of farm managers' and agronomists' work nature, involving continuous process management even outside the office, demanding mobile access to critical process information anytime, anywhere—all within a securely designed and implemented environment

\subsection{Platform Conceptual Architecture}

The SUSTAINABLE platform delivers a suite of core functionalities designed to enhance farm management. These include the end-to-end design, execution and monitoring of agricultural workflows, as well as integrated access to essential third-party tools and information sources. Based on stakeholder surveys, these features are tailored to support daily farm routines and are delivered through a secure, multilingual web interface that prioritizes user experience.

\begin{figure}[ht]
    \includegraphics[width=\columnwidth]{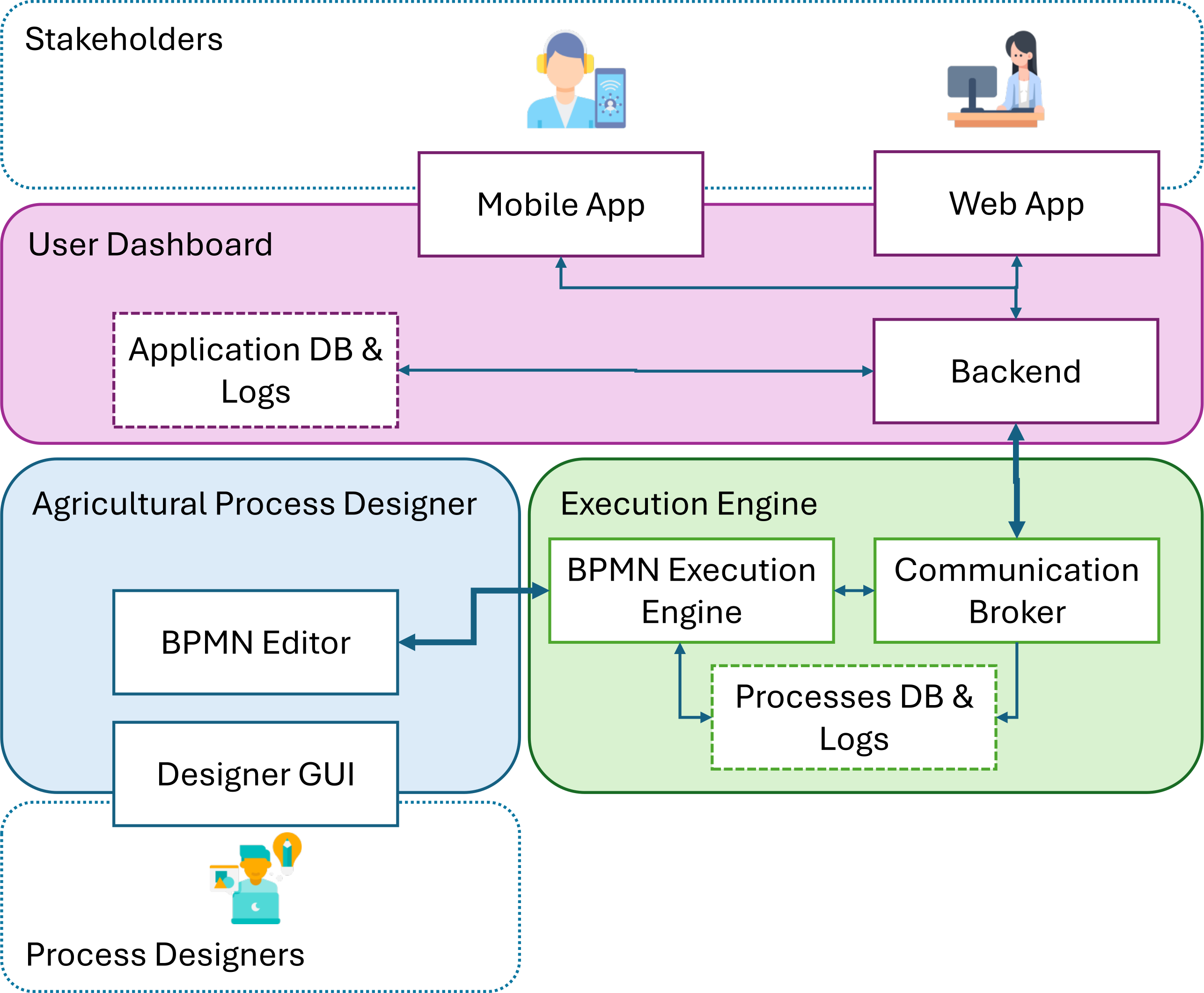}
    \caption{Conceptual Architecture. (\textit{This image has been designed using resources from Flaticon.com})}
    \label{fig:architecture}
\end{figure}

To deliver these capabilities, the platform is engineered as a unified system for managing a farm's work plan. Its architecture consists of three discrete, interconnected subsystems, which are themselves composed of smaller, specialized components. As illustrated in Fig. \ref{fig:architecture}, this modular design ensures the necessary flexibility and scalability to manage diverse agricultural processes. The three core subsystems are the \textit{Agricultural Processes Designer}, the \textit{Execution Engine}, and the \textit{User Dashboard}.

The \textit{Agricultural Processes Designer} is a tool that simplifies farming process design targeting farm managers and agronomists. It provides users with a simple and intuitive GUI that helps them design agriculture-related processes or transform existing ones. Process design is based on a formal description language, the Business Process Model and Notation (BPMN) \cite{bpmnspec:online}, a design based language adapted for the definition of business processes.  The process includes sequence of the actions to be taken, actors to perform actions, and  information to be collected and exchanged during task execution.  The output of the modeler is a valid BPMN file in XML format ready to be executed in the platform's execution engine, tailored to the needs of each farm. 

This BPMN file is then processed by the \textit{Execution Engine} subsystem; its heart being the BPMN execution engine module.  Moreover, the BPMN execution engine subsystem automates tasks that require constant monitoring, e.g detection of values exceeding temperature or humidity thresholds. BPMN execution engine is integrated with the User Dashboard through  a Communication Broker, which acts as an intermediate between all other components orchestrating information exchange. 

\textit{User Dashboard } corresponds to the presentation layer that notifies platform end users (farm managers, field workers etc.) displaying necessary information for carrying out needed tasks in a timely fashion. The User Dashboard consists of a back-end that is connected to the Execution Engine and feeds with the appropriate information two front-end applications. These two front-ends cater interaction of the system with its users, through two identical Graphical User Interfaces, one supporting web browsers, and a second supporting mobile devices, such as smartphones or tablets.

\subsection{Integration with External Agricultural Data Systems}

The goal of the platform is to allow all actors involved in the complex and diverse farm management procedures improve their efficiency and to assist them in making better decisions by:
\begin{itemize}
    \item unifying all the  sources of information,
    \item accelerating the dissemination of task assignments and field data,
    \item simplifying access to resources and documents describing the appropriate course of action.
\end{itemize}
Moreover, the process modeling utilizing a formal language such as BMPN helps clearly define the roles of all the involved actors, including farmers and external experts and services, such as:
\begin{itemize}
    \item \textbf{Farm Managers}: The farm management or owners responsible for defining and overseeing agricultural processes at the enterprise level.
    \item \textbf{Agronomists}: Agronomists are responsible for designing and monitoring  the agricultural tasks.
    \item \textbf{Field Workers}: This group includes all personnel and equipment operators who carry out the day-to-day tasks in the field.
    \item \textbf{Digital Platforms}: Platforms used to communicate tasks and progress to the team.
    \item \textbf{External Data Providers}: Third-party systems that supply supplemental data to support the decision-making process (e.g., weather services, satellite imagery platforms).
\end{itemize}

A core design principle of the SUSTAINABLE platform is interoperability with third-party systems. The platform addresses the challenge of heterogeneous data by first standardizing all inputs---from sensor readings to management directives---into a consistent internal model. This unified data is then made available to external systems via a secure REST API that utilizes the standard JSON data format. This technical stack was selected for its effectiveness in representing complex data structures and its status as a de facto standard for web-based data exchange.

This architecture enables the platform to act as a central hub, providing a single, standardized data stream to control various farm management software and machinery. The benefits of this approach are significant: it simplifies operations by eliminating the need to manage multiple disparate systems, improves technical reliability, and guarantees data consistency across all integrated platforms, which is crucial for modern farm management.

\subsection{External Agricultural Data Sources}

The SUSTAINABLE Platform needs to address the everyday operation of  stakeholders in the agricultural sector. In this context, the proposed platform is one-stop-point of integration of existing systems and sources of information currently used by farm managers and agronomists. The integration with the existing information systems and smart agricultural platforms is achieved by invoking provided web-based APIs, or integrating third-party applications' user graphical interfaces (i,e. using iframes), or by supporting manual file uploading in cases that the two systems cannot directly communicate (i.e. legacy old technologies, manual labor, communication restrictions).

\subsection{User Dashboard}

The platform's Dashboard serves as the central user interface, designed to support real-time decision-making. It provides a clear, intuitive way to monitor the farm's operational plan, access critical information, and manage tasks. The Dashboard's functionality is delivered through a set of distinct modules, each corresponding to tasks defined and automated by the BPMN execution engine.

This modular architecture is fundamental to the platform's design, ensuring both flexibility for future expansion and scalability to accommodate new farming scenarios. This approach also enhances the user experience by enabling administrators to create personalized dashboards, presenting each user with only the modules relevant to their specific role.

Furthermore, the Dashboard enforces a robust role-based access control (RBAC) system. It manages the type of information users can view based on their assigned roles and permissions. Users can be assigned to groups that reflect the farm's operational hierarchy, creating a flexible system that supports granular, multi-level access to information.

The primary modules of the Dashboard are described below:

\begin{itemize}
    \item \textbf{Task List:} As the core module, the Task List informs users of their pending tasks according to the active work plan. It offers detailed task descriptions and allows users to signal task completion to the system. For tasks that can be performed digitally, the module presents interactive web forms to guide the user through the required actions.
    
    \item \textbf{Notifications:} This module alerts users to new task assignments and critical agronomic warnings, such as high pest risk or frost alerts. While alerts are automatically generated by the BPMN engine, users have the capability to forward them to external experts who may not be part of the original process. These experts can be managed in user-defined contact lists for quick access.
    
    \item \textbf{Process Monitor:} This module provides an overview of the status of all ongoing work plans. Based on their role in the farm's hierarchy, users can view the overall progress of a plan and see the completion status of individual tasks.
    
    \item \textbf{External Data Views:} This module functions as a customizable virtual monitoring center, enabling users to aggregate various third-party data streams (e.g., weather forecasts) into a single, unified view. Users can select from a curated list of publicly available systems to build their own personalized monitoring dashboard.
\end{itemize}

\subsection{Business Process Management Notation Execution Engine}

\begin{figure*}[t]
    \centering
    \includegraphics[width=\textwidth]{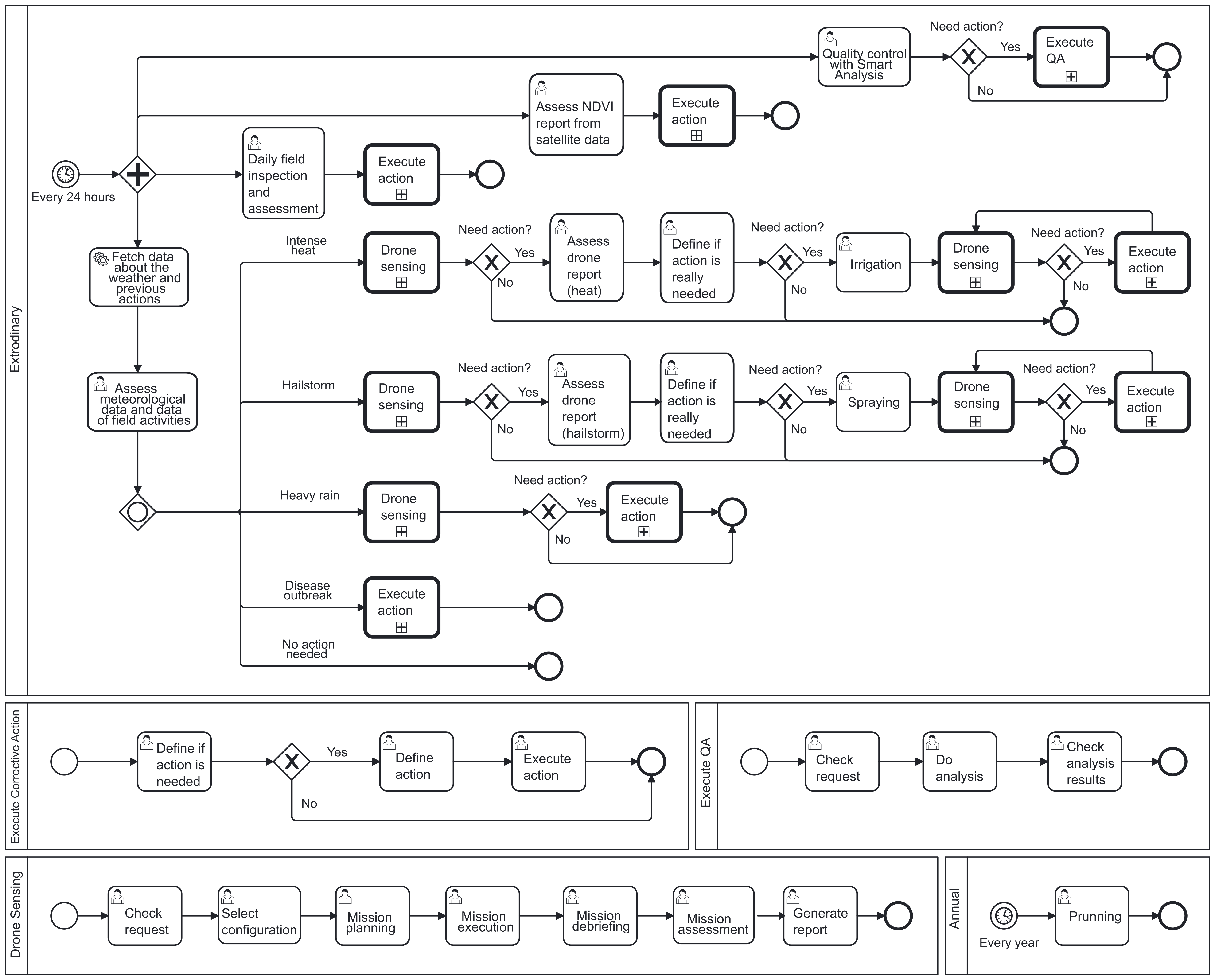}
    \caption{Orphanos Estate BPMN Process}
    \label{fig:camunda-bpmn}
\end{figure*}

This is the platform's core component for orchestrating the agricultural enterprise's operational plans through the design and execution of formal business processes. To model these workflows, we have adopted the Business Process Model and Notation 2.0 (BPMN 2.0) \cite{bpmnspec:online} standard. This specification was selected because it is the de facto industry solution, offering both straightforward translation into software components and complete independence from any particular technology stack.

Execution of these process models is managed by Camunda \cite{camunda:online}, a robust open-source engine for workflow and decision automation. While the conceptual design of a process can be performed with any BPMN 2.0-compliant software, the development of a deployable, executable artifact requires the use of the specialized Camunda Modeler. This tool is essential for linking the business model to the underlying technical implementation and deploying it to the engine.

From a technical standpoint, the Camunda Business Process Engine operates as a self-contained service exposed through a REST API. Its internal architecture is comprised of the following four primary subsystems:

\begin{itemize}
    \item \textbf{Public API:} Provides a command-driven interface that allows external client applications to engage with the engine.
    \item \textbf{BPMN 2.0 Core Engine:} The central processing unit, which includes a lightweight execution engine, a parser for converting BPMN 2.0 definitions into Java objects, and the necessary BPMN-specific logic.
    \item \textbf{Job Executor:} A dedicated component responsible for managing asynchronous operations and background tasks, such as time-triggered events.
    \item \textbf{Persistence Layer:} Manages the storage of all active process instance states in a relational database, ensuring process durability.
\end{itemize}

\section{Platform Use Case}

The SUSTAINABLE platform was evaluated by executing a use case scenario in a real-world farming environment. The scenario was conducted at a small, independent Greek winery that cultivates several native grape varieties \cite{orfanos:online} and involved modeling its daily and annual operational processes. These agricultural processes are supported by several external data systems. The Macro Expert Kft platform supplied weather, soil, and pest infection predictions, while Graniot \cite{graniot:online} provided processed satellite imagery. When satellite data was insufficient, GAIA Robotics \cite{gaiarobotics:online} gathered supplemental data using drones. Finally, DNA Phone \cite{dnaphone:online} performed on-demand smart quality control (QC) with its portable devices. Following a thorough analysis, the vineyard's complete operational workflow was documented in a BPMN diagram (Figure \ref{fig:camunda-bpmn}). Additionally, the necessary API callers were developed to enable communication with the third-party applications.

The procedure consists of four basic tasks executed in daily basis, and one task executed in annual basis. The first step in the modeling methodology is to highlight the main stakeholders that participate in the process. The identified stakeholders are: a) the viticulturist, the participant that manages all possible scenarios for the vineyard and decides which action is the best to implement, b) the field workers, c) Drone operator, managing drone sensing in the field, and d) QC device user. 

The next step is the definition of the process. Annually, the main task is the pruning of the plants. Daily, the main actions are: 
\begin{itemize}
    \item The definition of the need of smart analysis for quality control.
    \item The analysis of satellite images for exporting NDVI -related indexes report.
    \item On-site assessment of the plants from the viticulturist.
    \item Fetching daily weather forecast and positive disease outbreak probability warnings.
\end{itemize}
The weather forecast is of high importance as the viticulturist makes specific decisions for the vineyard after considering the weather of that specific day (whether intense heat is expected, hailstorm, heavy rain). The daily weather forecast is retrieved through a script which communicates with a weather API. In case of expected intense heat, a scenario triggered when expected temperature at the vineyard coordinates is above 35 degrees of Celsius, the viticulturist considers specific actions that should be actualized. The same thing happens with the expectation of hailstorm, heavy rainfall (a scenario which is triggered when total precipitation is above 6 mm), and a true disease outbreak probability warning. As depicted in Figure \ref{fig:camunda-bpmn}, the decisions and the tasks that are actualized after the trigger of a weather event vary. In any case, the viticulturist assesses weather conditions of the last 5 days, data which are retrieved from IoT sensing equipment, and the last times that irrigation and spraying were performed in the field. Furthermore, depending on the weather event that was triggered, the viticulturist may consider the implementation of drone sensing, to make irrigation or spraying in the field, and also define and execute a specific action after analyzing the aforementioned data (a task which is modeled by a sub-process). Furthermore, to evaluate indices like NDVI, the platform generates a color-coded map for the user. Each field is colored based on its calculated index value from the satellite imagery analysis. This visualization allows users to rapidly assess the condition of their own and neighboring fields and clearly delineates the precise boundaries of each parcel, which can otherwise be ambiguous.

In the BPMN model of this use case, three tasks are implemented as sub-processes. The first is the execution of smart analysis for quality control, while the second one is the drone sensing. In the last, drone operator follows a couple of different steps (like selecting configuration for the drone, making the mission planning debriefing, and assessment) before submitting the final report, which is read by the viticulturist. The final sub-process provides the definition of the execution of any specific action requested by the viticulturist to the field workers. This sub-process is triggered in any case of any weather event, in order to finalize the daily needs and actions that must be implemented in the vineyard.

The final BPMN was imported in the SUSTAINABLE's platform execution engine and the scenario was executed daily on May 2025. During pilot execution, the integration of the data and functionalities from external systems was successful with minor problems. The pilot users have easily understand the operation of the Dashboard, mainly due to its simplicity, managing to carry out all the scheduled task successfully. 

\section{Discussion and Conclusion}
The integration of multiple smart agriculture tools within one coherent platform improves decision-making, enables proactive interventions, and promotes sustainability. Challenges remain around data standardization, system interoperability, and scaling to broader crop types.

A core strength of the SUSTAINABLE platform lies in its ability to guide stakeholders through well-defined agricultural workflows, translating complex data into clear, executable tasks. For instance, agronomists can use processed satellite imagery and soil sensor data to recommend optimized fertilization schedules, while farmers receive timely alerts on irrigation needs or pest risks based on predictive analytics. The system also supports post-harvest quality control, with smart analyzers providing chemical composition reports that help producers meet food safety standards. Rather than simply merging datasets, the platform synthesizes information from existing solutions into practical recommendations, ensuring seamless adoption within daily farming operations. By maintaining a human-centric approach to automation, SUSTAINABLE bridges the gap between Agriculture 4.0 technologies and on-the-ground implementation, fostering more efficient, sustainable and stakeholder-driven farming practices.


SUSTAINABLE represents a step forward in smart viticulture by merging satellite analytics, IoT, and workflow orchestration into a unified platform. Future development of the platform is planned along several key trajectories. The first involves enhancing the core architecture by expanding its integration capabilities with a wider range of IoT sensors and autonomous agricultural machinery. To support this effort, a generalized API middleware will be developed. This middleware will abstract the complexities of individual integrations, providing a more scalable and standardized framework compared to the current approach of creating bespoke modules for each third-party service. A second area of development will be the introduction of an advanced analytics layer. This will involve the implementation of machine learning models designed to provide predictive insights for critical aspects of crop management and operational decision-making. Finally, future work will involve extensive user experience (UX) research and longitudinal case studies to validate the platform's real-world effectiveness and extend the platform's functionalities to additional stakeholders within the modern agricultural ecosystem. 

\bibliographystyle{IEEEtran}
\bibliography{sample}

@article{rose2018agriculture,
  title={Agriculture 4.0: Broadening responsible innovation in an era of smart farming},
  author={Rose, David Christian and Chilvers, Jason},
  journal={Frontiers in Sustainable Food Systems},
  volume={2},
  pages={87},
  year={2018},
  publisher={Frontiers Media SA}
}

@article{vos2019global,
  title={Global trends and challenges to food and agriculture into the 21st century},
  author={Vos, Rob and Bell{\`u}, Lorenzo Giovanni},
  journal={Sustainable Food and Agriculture},
  pages={11--30},
  year={2019},
  publisher={Elsevier}
}

@article{falcon2022rethinking,
  title={Rethinking global food demand for 2050},
  author={Falcon, Walter P and Naylor, Rosamond L and Shankar, Nikhil D},
  journal={Population and Development Review},
  volume={48},
  number={4},
  pages={921--957},
  year={2022},
  publisher={Wiley Online Library}
}

@article{bhatti2024global,
  title={Global production patterns: Understanding the relationship between greenhouse gas emissions, agriculture greening and climate variability},
  author={Bhatti, Uzair Aslam and Bhatti, Mughair Aslam and Tang, Hao and Syam, MS and Awwad, Emad Mahrous and Sharaf, Mohamed and Ghadi, Yazeed Yasin},
  journal={Environmental Research},
  volume={245},
  pages={118049},
  year={2024},
  publisher={Elsevier}
}

@article{barrett2021overcoming,
  title={Overcoming global food security challenges through science and solidarity},
  author={Barrett, Christopher B},
  journal={American Journal of Agricultural Economics},
  volume={103},
  number={2},
  pages={422--447},
  year={2021},
  publisher={Wiley Online Library}
}

@article{batra2024industrial,
  title={Industrial revolution and smart farming: a critical analysis of research components in Industry 4.0},
  author={Batra, Isha and Sharma, Chetan and Malik, Arun and Sharma, Shamneesh and Kaswan, Mahender Singh and Garza-Reyes, Jose Arturo},
  journal={The TQM Journal},
  year={2024},
  publisher={Emerald Publishing Limited}
}

@article{ahmad2024ai,
  title={AI can empower agriculture for global food security: challenges and prospects in developing nations},
  author={Ahmad, Ali and Liew, Anderson XW and Venturini, Francesca and Kalogeras, Athanasios and Candiani, Alessandro and Di Benedetto, Giacomo and Ajibola, Segun and Cartujo, Pedro and Romero, Pablo and Lykoudi, Aspasia and others},
  journal={Frontiers in Artificial Intelligence},
  volume={7},
  pages={1328530},
  year={2024},
  publisher={Frontiers Media SA}
}

@article{ojha2021internet,
  title={Internet of things for agricultural applications: The state of the art},
  author={Ojha, Tamoghna and Misra, Sudip and Raghuwanshi, Narendra Singh},
  journal={IEEE Internet of Things Journal},
  volume={8},
  number={14},
  pages={10973--10997},
  year={2021},
  publisher={IEEE}
}

@article{yepez2023mobile,
  title={Mobile robotics in smart farming: current trends and applications},
  author={Y{\'e}pez-Ponce, Dar{\'\i}o Fernando and Salcedo, Jos{\'e} Vicente and Rosero-Montalvo, Pa{\'u}l D and Sanchis, Javier},
  journal={Frontiers in Artificial Intelligence},
  volume={6},
  pages={1213330},
  year={2023},
  publisher={Frontiers Media SA}
}

@article{chergui2022data,
  title={Data analytics for crop management: a big data view},
  author={Chergui, Nabila and Kechadi, Mohand Tahar},
  journal={Journal of Big Data},
  volume={9},
  number={1},
  pages={123},
  year={2022},
  publisher={Springer}
}

@misc{xfarm,
  title={xFarm Smart Farming Platform},
  year={2024},
  note={xFarm Technologies},
  howpublished={\url{https://www.xfarm.ag/}}
}

@misc{cropx,
  title={Soil Intelligence Platform},
  year={2024},
  note={CropX Technologies},
  howpublished={\url{https://www.cropx.com/}}
}

@misc{fieldview,
  title={Climate FieldView: Digital Farming Platform},
  year={2024},
  note={Bayer},
  howpublished={\url{https://climatefieldview.com/}}
}

@misc{onesoil,
  title={Precision Farming with Satellite Maps},
  year={2024},
  note={OneSoil},
  howpublished={\url{https://onesoil.ai/}}
}

@article{kulatunga2024,
  title={Machine Learning for Dynamic Management Zone in Smart Farming},
  author={Kulatunga, Chathura and Dhelim, Shaoxuan and Kechadi, Tahar},
  journal={arXiv preprint arXiv:2408.00789},
  year={2024}
}

@article{mazzia2020,
  title={UAV and ML Based Refinement of Satellite NDVI for Precision Agriculture},
  author={Mazzia, V. and others},
  journal={arXiv preprint arXiv:2004.14421},
  year={2020}
}

@article{kumar2024,
  title={A Comprehensive Review on Smart and Sustainable Agriculture using IoT Technologies},
  author={Kumar, V. and others},
  journal={Elsevier},
  year={2024}
}

@misc{sustainable,
  title={Stop running, stop and start using our knowledge to be reachable},
  year={2022},
  note={sustainable},
  howpublished={https://projectsustainable.eu/}}

@misc{camunda:online,
  author       = {{Camunda}},
  title        = {Camunda: The Universal Process Orchestrator},
  howpublished = {\url{https://camunda.com/}},
  year         = {2025},
  note         = {Accessed on June 18, 2025}
}

@misc{bpmnspec:online,
  author = {{Object Management Group}},
  title  = {Business Process Model and Notation (BPMN) Version 2.0.2},
  year   = {2013},
  url    = {https://www.omg.org/spec/BPMN/2.0.2/},
  note   = {Accessed on June 18, 2025}
}

@misc{gaiarobotics:online,
  author       = {{GAIA Robotics}},
  title        = {GAIA Robotics},
  howpublished = {\url{https://www.gaiarobotics.gr/}},
  year         = {2025},
  note         = {Accessed on June 18, 2025}
}

@misc{dnaphone:online,
  author       = {{DNA Phone}},
  title        = {DNA Phone},
  howpublished = {\url{https://www.dnaphone.it/}},
  year         = {2025},
  note         = {Accessed on June 18, 2025}
}

@misc{graniot:online,
  author       = {{GRANIOT}},
  title        = {GRANIOT},
  howpublished = {\url{https://web.graniot.com/}},
  year         = {2025},
  note         = {Accessed on June 18, 2025}
}

@article{alexopoulos2023complementary,
  title={Complementary use of ground-based proximal sensing and airborne/spaceborne remote sensing techniques in precision agriculture: A systematic review},
  author={Alexopoulos, Angelos and Koutras, Konstantinos and Ali, Sihem Ben and Puccio, Stefano and Carella, Alessandro and Ottaviano, Roberta and Kalogeras, Athanasios},
  journal={Agronomy},
  volume={13},
  number={7},
  pages={1942},
  year={2023},
  publisher={MDPI}
}

@misc{orfanos:online,
  author       = {{Orfanos Estate}},
  title        = {Orfanos Estate},
  howpublished = {\url{https://ktimaorfanou.gr/}},
  year         = {2025},
  note         = {Accessed on June 18, 2025}
}

\end{document}